\documentclass[11pt]{article}

\usepackage[]{acl}

\usepackage{times}
\usepackage{latexsym}
 \usepackage{booktabs}
\usepackage[T1]{fontenc}

\usepackage[utf8]{inputenc}

\usepackage{microtype}

\usepackage{inconsolata}

\usepackage{graphicx}

\def\OURAGENT{Asclepius}

\title{Asclepius: An Adaptive Harness for Long-Horizon Clinical Agents}

\author{
  \textbf{Grace Chang Yuan\textsuperscript{1}},
  \textbf{Xiaoman Zhang\textsuperscript{2}},
  \textbf{Sung Eun Kim\textsuperscript{2}},
  \textbf{Luyang Luo\textsuperscript{2}},
  \textbf{Pranav Rajpurkar\textsuperscript{2}}
\\
\\
  \textsuperscript{1}Massachusetts Institute of Technology, Cambridge, MA
\\
  \textsuperscript{2}Department of Biomedical Informatics, Harvard Medical School, Boston, MA
\\
  \small{
    \textbf{Correspondence:} \href{mailto:yuangc@mit.edu}{yuangc@mit.edu},
    \href{mailto:pranav_rajpurkar@hms.harvard.edu}{pranav\_rajpurkar@hms.harvard.edu}
  }
}

\begin{document}
\maketitle
\begin{abstract}

LLM agents are predominantly benchmarked on short, single-task trajectories,
yet real deployments run for hours under contention, surfacing a different
class of failures. We use the Clinical Environment Simulator (CES), in which
an agent manages an entire emergency-department shift under continuous time
and resource pressure, as a testbed: long-horizon execution failures manifest
measurably in a single rollout under structured, multi-dimensional grading.
On CES, current agents reach the correct diagnosis in most cases yet fail to
deliver complete and timely critical actions, revealing an \emph{execution
gap}. We attribute this gap to three long-horizon failure modes, each
operationalized as a per-trace counter: instruction-adherence drift, treatment
incompleteness, and a severity-equity gap in timeliness. We then introduce
\textsc{Asclepius}, an adaptive agent scaffolding with a self-evolving harness
that rewrites the operating manual between shifts from trace-level feedback,
an externalized clinical skills library for high-stakes regimen knowledge, and
three isolated subagents that partition per-turn decisions across the patient
queue. On held-out batches never observed during harness evolution,
\textsc{Asclepius} improves critical-action correctness by 22\%
($p = 0.024$) over a strong baseline agent framework while preserving
diagnostic accuracy, with consistent gains across five LLM judges from three
model families; on the full ten-batch set, improvements reach 25\% on critical
actions and 13\% on timeliness. The three failure modes form a \emph{coupled
bottleneck}: decisive reductions appear only when all three components act
together.
\end{abstract}
\section{Introduction}
\label{sec:introduction}

\begin{figure*}[t]
    \centering
    \includegraphics[width=\textwidth]{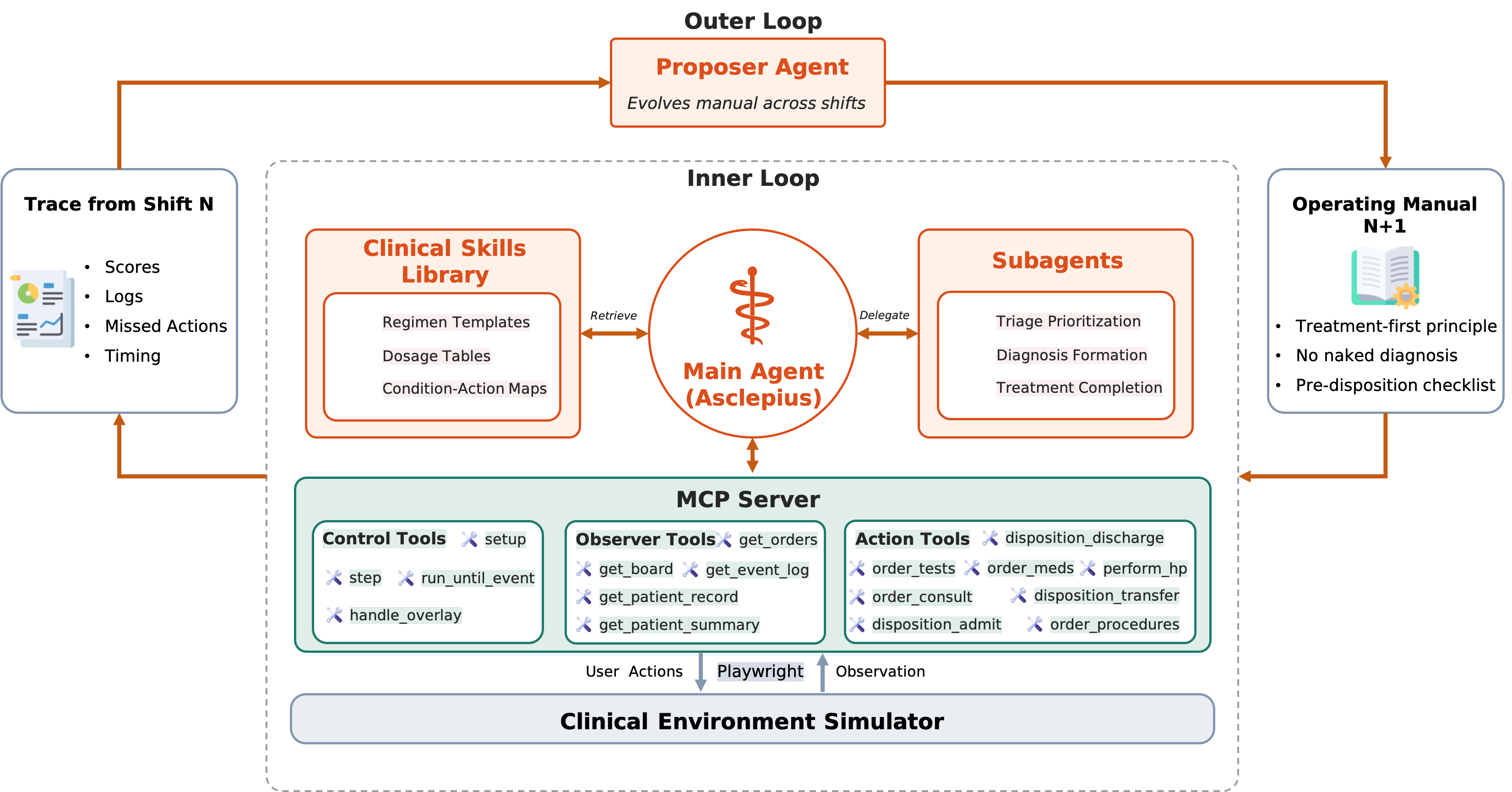}
    \caption{
        \textbf{Overview of \OURAGENT{}.} Two coupled loops wrap a baseline agent. Within a shift (inner loop), the Main Agent uses an MCP tool interface, a clinical skills library, and three isolated subagents. Across shifts (outer loop), a proposer agent reads the shift trace and rewrites the operating manual for the next shift.
    }
    \label{fig:overview}
\end{figure*}


\noindent\textbf{From Short-Horizon to Long-Horizon Agent Deployment.}
Progress in LLM agents has been driven largely by short-horizon, single-task benchmarks where success is scored at the end of a brief trajectory. Yet real deployments demand sustained competent execution across hours of interleaved tasks under contention, where a different class of failures emerges that is invisible to per-task accuracy but dominates end-of-shift performance. We use the Clinical Environment Simulator (CES)~\citep{luo2026clinical} as our testbed: a long-horizon emergency-department shift in which patients arrive on staggered schedules with hidden deterioration thresholds, vitals evolve continuously, and the agent must interleave diagnosis with time-sensitive interventions under contention for limited beds and staff. Performance is graded along four dimensions: diagnosis, critical actions, timeliness, and disposition.

\vspace{3pt} \noindent\textbf{Gaps in Current Agents.} 
We evaluate four configurations representative of current agent practice on CES: three raw frontier-model agents (Codex~\citep{openai2025codex}, Gemini~\citep{google2025gemini}, and Claude Code~\citep{anthropic2025claudecode}) and a stronger baseline that wraps Claude Code in an MCP-based clinical tool interface with a hand-crafted operating manual. 
The tool-using configuration substantially outperforms the raw agents, yet a striking asymmetry persists across all four: diagnostic accuracy is high (4.39/5) while critical-action scores remain below 3.0/5 (2.94/5) and timeliness while higher, still trails at 3.34/5. We call this the \emph{execution gap}: the failure to translate correct diagnosis into complete and timely action under multi-patient load.

\vspace{3pt} \noindent\textbf{Failure Modes.}
We identify three failure modes that underlie this execution gap, each operationalized with one per-trace counter (Section~\ref{sec:results-failure-modes}, Figure~\ref{fig:failure-modes}).
\textbf{(i)}~\emph{Instruction-adherence drift}: a hand-crafted operating manual that works at hour zero decays in adherence as the shift lengthens and clinical trajectories diversify, a runtime problem that hand-tuning cannot scale to address; measured by the \emph{Drift} counter (early$\to$late rise in naked-disposition rate).
\textbf{(ii)}~\emph{Treatment incompleteness}: even with a correct diagnosis, the agent frequently orders zero disease-specific treatment, partial multi-drug regimens, or familiar defaults instead of guideline-specified drugs; measured by the \emph{Incomplete} counter (zero-treatment rate).
\textbf{(iii)}~\emph{Severity-equity gap}: as the queue widens, the agent's attention collapses onto recent and easier cases, leaving harder and severe patients with systematically delayed care; measured by the \emph{Neglect} counter (timeliness gap between non-severe and severe patients).

\vspace{3pt} \noindent\textbf{\OURAGENT{}.} We propose \textsc{Asclepius}, an adaptive agent scaffolding with three components, each targeting one of the failure modes above.
\textbf{(i)}~A \textbf{self-evolving harness} treats the operating manual as a learnable natural-language artifact: a proposer agent reads trace-level feedback (scores, actions, missed treatments) from completed shifts and rewrites the manual between iterations.
\textbf{(ii)}~A curated \textbf{clinical skills library} externalizes procedural treatment-selection knowledge (regimen templates, dosage tables, condition-to-action mappings) as an on-demand reference.
\textbf{(iii)}~Three \textbf{isolated subagents} (triage prioritization, diagnosis formation, treatment completion) partition per-turn decisions into focused context windows.


\vspace{3pt} \noindent\textbf{Results.} We evaluate Asclepius on ten patient batches in CES (six
search and four held out from manual evolution). Pooled across all ten
batches, Asclepius delivers gains on critical actions ($+25\%$,
$p < 0.001$) and timeliness ($+13\%$, $p < 0.01$) while preserving
diagnostic accuracy. On the four held-out batches alone ($n = 48$ patients),
the critical-action gain remains significant ($+0.63$, $p = 0.024$);
overall ($+0.19$) and timeliness ($+0.31$) remain positive but not
significant. Five judges from three model families reproduce the same
pattern, so the improvement is not an artifact of the original grader. Component ablations show that no single component closes the execution gap on its own. The three failures behave as a
\emph{coupled bottleneck}: no single component reduces all three per-trace
counters, and the largest joint reduction appears only when the self-evolving
manual, the skills library, and the isolated subagents act together.

\section{Related Work}
\label{sec:relatedwork}

\subsection{Clinical LLM Agents and Evaluation}
\label{sec:rw-clinical-agents}
Evaluation of LLMs in medicine has progressed through three regimes. Early work used static, single-question benchmarks drawn from licensing exams and curated vignettes, where models answer one case in isolation. A second wave introduced multi-turn diagnostic agents that interview a single patient or reason over a single case across multiple steps: AMIE conducts conversational history-taking~\cite{tu2025conversational, mcduff2025differential}, MAI-DxO operationalizes diagnosis as sequential test ordering~\cite{nori2025sequential}, and benchmarks such as AgentClinic and MedAgentBench evaluate agents in simulated clinical environments~\cite{schmidgall2024agentclinic, jiang2025medagentbench}. Multi-agent variants extend this paradigm by simulating entire hospitals~\cite{li2024agenthospital, fan2025aihospital} or expanding tool use into therapeutic and rare-disease reasoning~\cite{gao2025txagent, zhao2026agentic}. These systems remain single-patient and untimed, with evaluation dominated by the diagnosis dimension. Recent work argues that clinical AI evaluation must move toward continuous, multi-patient simulators that grade time-sensitive treatment actions alongside diagnosis~\cite{luo2026clinical}. Our work targets the resulting long-horizon execution gap in such environments.

\subsection{Agent Harnesses}
\label{sec:rw-harness}
We use \emph{harness} to refer to the non-parametric scaffolding around an LLM, including prompts, tools, control flow, and inter-agent coordination, that determines how a fixed model executes a task. The dominant pattern equips agents with external tools and skill libraries: ReAct interleaves tool calls with chain-of-thought reasoning~\cite{yao2023react}, Toolformer fine-tunes the model to invoke APIs autonomously~\cite{schick2023toolformer}, and Voyager introduces a learned, growable skill library for embodied agents~\cite{wang2024voyager}. These approaches treat skills as task-level procedures discovered or invoked during execution, and rely on the underlying model's parametric memory for domain knowledge. Our skills component takes a different design point: rather than learning skills end-to-end or growing them online, we curate a small, condition-indexed library of high-stakes clinical knowledge (regimen templates, dosage tables, and condition-to-action mappings) explicitly targeted at the incomplete-treatment-selection failure mode identified in rollout traces.

\subsection{Prompt Optimization and Self-Evolving Harnesses}
\label{sec:rw-meta-harness}
A growing line of work treats the prompt as a learnable artifact. APE searches over instructions with an LLM as the search operator~\cite{zhou2023ape}; OPRO frames the LLM as a black-box optimizer over textual prompts~\cite{yang2024opro}; DSPy jointly optimizes prompts and program structure against task metrics~\cite{khattab2024dspy}. These methods operate on single-prompt input/output pairs and optimize against a scalar metric. A recent survey of agent externalization identifies \emph{self-evolving harnesses}, in which the orchestration logic itself is rewritten between rollouts, as an emerging direction~\cite{zhou2026externalization}. Its closest concrete realization is Meta-Harness~\cite{lee2026metaharness}, which runs an agentic proposer with filesystem access to the code, scores, and execution traces of all prior candidates. \OURAGENT{} shares this proposer-over-traces design, but evolves a natural-language operating \emph{manual} rather than harness code, over multi-hour, multi-patient clinical rollouts with structured per-dimension feedback (diagnosis, critical actions, timeliness, disposition) rather than a scalar reward. In the clinical domain, EvoClinician \citep{he2026evoclinician} performs
test-time evolutionary learning to adapt diagnostic strategy within a
multi-turn encounter with a single patient. Asclepius is complementary along both axes: the unit of adaptation is the cross-shift operating manual rather
than the within-encounter policy, and the target is execution under
multi-patient load rather than
diagnostic accuracy, with the underlying model and diagnosis pipeline held fixed.
HORIZON~\cite{wang2026long} characterizes long-horizon agent failure across four domains by assigning each failed trajectory a single primary label from a seven-category taxonomy; our characterization targets a complementary regime, operationalizing three failure modes as continuous per-trace counters under multi-objective, continuously-graded execution and finding them \emph{coupled} rather than orthogonal.

\section{Preliminaries}

\subsection{Clinical Environment Simulator}
\label{sec:ces}

We evaluate on the Clinical Environment Simulator (CES)~\cite{luo2026clinical}, a turn-based discrete-event simulator of a multi-hour, multi-patient emergency-department shift. Patients arrive on staggered schedules with hidden deterioration thresholds; vitals evolve continuously, ordered tests return on simulated delays, and complications develop over time. The agent interacts with the simulator through an EHR-style interface and is graded by an LLM judge along four dimensions on a 1--5 scale: diagnosis, critical actions, timeliness, and disposition. Full platform mechanics are in Appendix~\ref{app:simulator}; rubric details are in Appendix~\ref{app:evaluation}.

\subsection{Baseline Agents}
\label{sec:ces-base}
We compare against two baseline configurations.

\vspace{2pt}\noindent\textbf{Single-LLM agent.}
A frontier LLM is given a minimal task description and direct access to the simulator interface, with no hand-crafted operating manual, no skill resources, and no subagents. We report this configuration with three frontier models: Codex, Gemini, and Claude Code. Claude Code appears in both baselines: as a raw model here, and as the underlying model of the framework below.
 
\vspace{2pt}\noindent\textbf{Baseline agent framework.}
A stronger baseline wraps Claude Code in an MCP-based clinical tool interface and a hand-crafted operating manual. The interface exposes \emph{observation} tools (EHR content), \emph{action} tools (orders, examinations, dispositions), and \emph{control} tools (simulator lifecycle); natural-language order strings are resolved to catalog entries through a three-stage matching pipeline. The operating manual instructs the agent on triage, parallel workup, and disposition. The agent maintains the full multi-patient state in a single context window and has no specialized knowledge resources beyond parametric memory. This configuration serves as the starting point for \OURAGENT{}; tool-by-tool details are in Appendix~\ref{app:baseline}.

\section{\OURAGENT{}}
\label{sec:method}

Figure~\ref{fig:overview} shows the architecture as two coupled loops: an \emph{outer loop} that rewrites the operating manual between shifts, and an \emph{inner loop} that runs within each shift and consists of a skills library and three subagents. 

\subsection{Outer Loop: Self-Evolving Harness}
\label{sec:method-harness}
 
Let $M_t$ denote the operating manual at iteration $t \in \{0, 1, 2, \dots\}$ and $P$ a proposer agent (a separate agent with no access to the simulator, the underlying patient scenario definitions, or the held-out batches). At iteration $t$, $P$ receives a filesystem $F_t$ containing the full artifacts of all candidates produced so far $\{M_0, \dots, M_t\}$: each candidate's manual, aggregate, and per-batch scores, per-patient judge rows, per-turn action records, and LLM call audit logs. $P$ navigates $F_t$ to inspect prior manuals and traces, identifies the current best-scoring manual as the editing base, and outputs a revised manual $M_{t+1}$. We evaluate $M_{t+1}$ on the search batches, append its scores to the candidate record, and after $N$ iterations select the best-scoring manual on the search batches as $M^*$.
 
The proposer's system prompt requires it to: (i)~base each edit on a failure pattern observed in $\geq 3$ patients across multiple batches, (ii)~avoid patient- or batch-specific hard-coded knowledge, and (iii)~produce substantive behavioral edits rather than cosmetic rewording. Constraints (i)--(ii) guard against overfitting to the search batches; (iii) prevent wasted iterations.
 
\vspace{2pt}\noindent\textbf{Search and held-out validation.}
We partition the patient batches into six search batches (visible to the proposer) and four held-out batches (never observed during iteration). The held-out split tests whether evolved manuals generalize beyond the search trajectories. We use Claude Code with Claude Opus 4.6~\citep{anthropic2026claudeopus46} as $P$ and run $N=10$ iterations, selecting the manual with the best search-batch overall score as $M^*$ ($M_7$, denoted v7 in Section~\ref{sec:results}).
 
\subsection{Inner Loop: Skills Library}
\label{sec:method-skills}
 
The skills library provides on-demand reference modules that the Main Agent retrieves during treatment selection. The \emph{clinical decision guide} contains a presentation-to-management table specifying disease-specific drugs, doses, routes, and non-pharmacologic interventions, together with a seven-category treatment-completeness framework. The \emph{clinical reasoning} module contains the treatment-first principle, the ``no naked diagnosis'' rule, a pre-disposition verification checklist, and a catalog of common failure patterns. A separate \emph{simulation mechanics} module covers turn structure, action lifecycle, and resource constraints. Skills are loaded automatically when their description matches the current context and appended to the working context as structured reference.
 
\subsection{Inner Loop: Subagents}
\label{sec:method-subagents}
 
We externalize three recurring per-turn decisions to subagents that run as separate LLM calls with isolated contexts: (i)~\emph{triage prioritization} selects the next patient to attend to; (ii)~\emph{diagnosis formation} synthesizes test results into a working diagnosis and disposition recommendation; (iii)~\emph{treatment completion} identifies missing tests, medications, and procedures for a diagnosed patient and ensures they are ordered.
 
Subagent prompts contain only content extracted from the hand-crafted
operating manual $M_0$ and are held fixed across all configurations; no
additional medical knowledge is introduced. Gains from subagent decomposition
therefore come from focused context per decision, structured-output
enforcement, and inter-subagent coordination, rather than from added
knowledge.
 
\vspace{2pt}\noindent\textbf{Advisory vs. acting variants.}
We evaluate two variants. In the \emph{advisory} variant, subagents have no MCP tool access; they receive a structured snapshot of simulator state and return a textual recommendation that the Main Agent parses and executes. In the \emph{acting} variant, subagents have direct MCP access within their scope (the triage subagent rooms patients, the diagnosis subagent commits dispositions and consults, the treatment subagent places missing orders); the Main Agent serves as a coordinator that selects which subagent to invoke at each turn. We use the acting variant as the default in Section~\ref{sec:results} and report the other as an ablation.
\section{Experimental Setup}
\label{sec:experiments}

\subsection{The Clinical Simulator}
\label{sec:exp-ces}

We evaluate on the Clinical Environment Simulator
(CES)~\citep{luo2026clinical}, a long-horizon multi-patient
emergency-department simulator in which an LLM agent manages a multi-hour
shift: new patients arrive throughout the shift following a stochastic
arrival process, existing patients evolve continuously (vitals shifting,
ordered tests returning on simulated delays, complications developing over
time), and the agent issues actions from a fixed set (laboratory or imaging
orders, medications, consults, dispositions) through an EHR tool interface.

Each shift is graded by an LLM judge with access to the gold patient
trajectory along four dimensions on a 1 to 5 scale: \emph{diagnosis},
\emph{critical actions}, \emph{timeliness}, and
\emph{disposition}~\citep{luo2026clinical}. Higher scores reflect closer
correspondence to expert-graded reference trajectories. We report
per-dimension means and the overall mean across all dimensions. Our primary
judge is GPT-4.1~\citep{openai2025gpt41}, held fixed across all
configurations and used for every table unless stated otherwise.

To test whether results depend on the choice of grader, we additionally
re-grade every patient under the baseline and full \OURAGENT{} configurations
(120 patients per configuration) with four further judges spanning three
model families: Claude Sonnet 4.5 \citep{anthropic2025claudesonnet45},
Claude Sonnet 5 \citep{anthropic2026claudesonnet5}, Gemini 3.1 Pro
\citep{google2026gemini31pro}, and GPT-5 \citep{openai2025gpt5}. Each
validation judge receives the identical rubric, answer keys, and trajectory
records as the primary judge; only the grading model changes. The component
ablations are graded by the primary judge alone. Results are reported in
Table~\ref{tab:multijudge}.

Full simulator mechanics, baseline agent configuration details, and
evaluation rubrics are provided in Appendices~\ref{app:simulator},
\ref{app:baseline}, and \ref{app:evaluation}, respectively.
\subsection{Patient Batches and Evaluation Protocol}
\label{sec:exp-protocol}

Each patient batch contains twelve patients with varying acuities, arrival times, and underlying diagnoses. We use ten batches in total, partitioned into six \emph{search} batches used during harness evolution and four \emph{held-out validation} batches reserved for final evaluation. The validation batches are sampled from the same distribution as the search batches but are never observed by the proposer agent. Each configuration is run once on every search and validation batch, and per-dimension scores are averaged across batches.

\subsection{Configurations}
\label{sec:exp-configs}

We compare two primary configurations and several ablations:
\begin{itemize}
    \item \textbf{Raw single-agent}: a frontier model invoked with only the simulator tool interface and a minimal task description, with no hand-crafted operating manual, no skills, no subagents, and no harness evolution. We report three frontier-model variants under this configuration: Codex, Gemini, and Claude Code.
    \item \textbf{Baseline agent configuration}: the strongest agent baseline we have on CES, comprising a single frontier LLM, the simulator tool interface, and a hand-crafted operating manual (Appendix~\ref{app:baseline}). We re-implement this configuration as our starting point and report it under identical conditions to \OURAGENT{}.
    \item \textbf{\OURAGENT{}}: our full system (Section~\ref{sec:method}), wrapping the baseline configuration with the self-evolving harness, skills, and subagents.
\end{itemize}
For component ablations we evaluate \OURAGENT{} variants that add the three components one at a time on top of the baseline configuration.

\subsection{Implementation Details}
\label{sec:exp-impl}

We use Claude Code, with Claude Opus~4.6~\citep{anthropic2026claudeopus46} as the underlying model, for both the
main agent and the proposer agent. The harness is evolved for ten iterations over the six search batches; we report the results on both search and validation. 

\begin{table*}[t]
\centering
\caption{\textbf{Main results across all ten patient batches.}
Mean scores (out of 5) for the three raw single-agent variants, the baseline agent framework, and \OURAGENT{} (harness + skills + acting subagents). Scores pool the six search batches, over which the operating manual was evolved, with the four held-out batches. Significance for the $\Delta$ row is from a patient-level mixed-effects model, BH-FDR corrected across all 20 system-by-dimension tests (*: $p<0.05$, **: $p<0.01$, ***: $p<0.001$, ns: not significant). Bold marks the best score in each column.}
\label{tab:main}
\small
\setlength{\tabcolsep}{12pt}
\begin{tabular}{lccccc}
\toprule
Configuration & Overall & Diagnosis & Critical Actions & Timeliness & Disposition \\
\midrule
Codex (raw)                  & 2.74 & 3.74 & 1.59 & 1.58 & 4.03 \\
Gemini (raw)                 & 2.06 & 2.47 & 1.43 & 1.42 & 2.93 \\
Claude Code (raw)            & 3.52 & 4.22 & 2.69 & 2.65 & 4.53 \\
\midrule
Baseline framework           & 3.80 & \textbf{4.39} & 2.94 & 3.34 & 4.52 \\
\OURAGENT{}                  & \textbf{4.10} & 4.38 & \textbf{3.67} & \textbf{3.79} & \textbf{4.54} \\
\midrule
$\Delta$ vs.\ baseline       & $+0.30$\,** & $-0.02$\,\textsuperscript{ns} & $+0.73$\,*** & $+0.45$\,** & $+0.03$\,\textsuperscript{ns} \\
\bottomrule
\end{tabular}
\end{table*}

\begin{table}[t]
\centering
\caption{\textbf{Held-out significance (batches 7--10, $n=48$).}
Mixed-effects mean differences of \OURAGENT{} vs.\ the baseline framework, fit on the held-out batches alone. Diagnosis and disposition are unchanged (both ns).}
\label{tab:heldout}
\small
\begin{tabular}{lcc}
\toprule
Dimension & $\Delta$ & $p$ \\
\midrule
Critical Actions & $+0.625$ & $0.024$\,* \\
Timeliness       & $+0.312$ & ns \\
Overall          & $+0.193$ & ns \\
\bottomrule
\end{tabular}
\end{table}

\begin{table*}[t]
\centering
\caption{\textbf{Multi-judge validation.}
$\Delta$ = \OURAGENT{} $-$ Baseline under each of the five judges. (*: $p<0.05$, **: $p<0.01$, ***: $p<0.001$, ns: not significant).}
\label{tab:multijudge}
\small
\setlength{\tabcolsep}{10pt}
\begin{tabular}{lccccc}
\toprule
Judge & $\Delta$ Overall & $\Delta$ Critical Actions & $\Delta$ Timeliness & $\Delta$ Diagnosis & $\Delta$ Disposition \\
\midrule
GPT-4.1 (primary) & $+0.30$\,**  & $+0.73$\,*** & $+0.45$\,**  & $-0.02$\,\textsuperscript{ns} & $+0.03$\,\textsuperscript{ns} \\
Claude Sonnet 4.5 & $+0.30$\,*   & $+0.65$\,*** & $+0.52$\,**  & $-0.04$\,\textsuperscript{ns} & $+0.09$\,\textsuperscript{ns} \\
Claude Sonnet 5   & $+0.36$\,**  & $+0.76$\,*** & $+0.72$\,*** & $-0.02$\,\textsuperscript{ns} & $-0.03$\,\textsuperscript{ns} \\
Gemini 3.1 Pro    & $+0.30$\,*   & $+0.67$\,*** & $+0.67$\,*** & $-0.02$\,\textsuperscript{ns} & $-0.12$\,\textsuperscript{ns} \\
GPT-5             & $+0.31$\,**  & $+0.65$\,*** & $+0.71$\,*** & $-0.05$\,\textsuperscript{ns} & $-0.09$\,\textsuperscript{ns} \\
\bottomrule
\end{tabular}
\end{table*}

\begin{figure*}[t]
    \centering
    \includegraphics[width=1\linewidth]{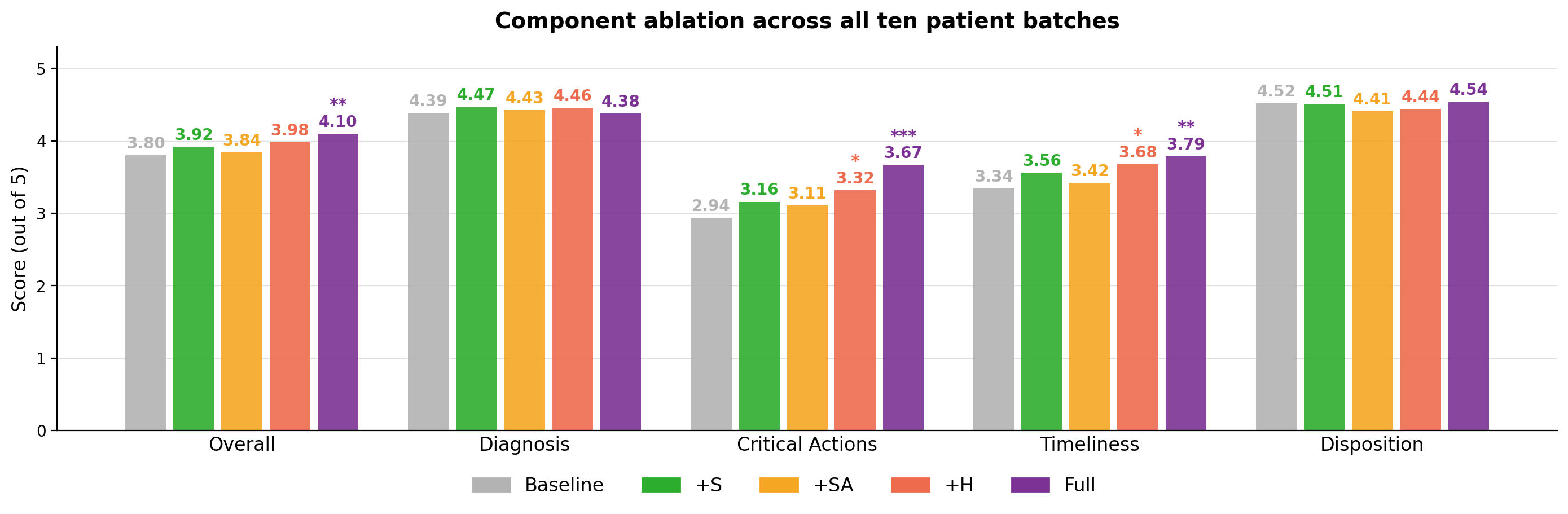}
    \caption{\textbf{Component ablation across all ten patient batches.} Per-dimension scores for the baseline framework, baseline~+~skills, baseline~+~subagents, baseline~+~harness, and \OURAGENT{} (Full). The baseline, $+S$, and $+SA$ use the hand-crafted manual $M_0$; $+H$ and full \OURAGENT{} use the evolved manual $M^*$.}
    \label{fig:ablation}
\end{figure*}

\section{Results}
\label{sec:results}

We report aggregate scores across all ten patient batches (120 patients total). Significance for each comparison against the baseline framework comes from a patient-level mixed-effects model with random effects for patient and batch, with $p$-values BH-FDR \cite{benjamini1995controlling} corrected across the 20 system-by-dimension tests. Search-vs-held-out generalization, per-batch breakdowns, and the full significance table are in Appendix~\ref{app:results-per-batch}.

\subsection{Main Results}
\label{sec:results-main}

Table~\ref{tab:main} compares \OURAGENT{} against three raw single-agent variants and the baseline agent framework. The raw single-agents span a wide range (Gemini 2.06 to Claude Code 3.52 overall), and even the strongest raw variant trails the baseline framework by 0.28 overall, attributing substantial value to the MCP tool interface and hand-crafted operating manual alone. The execution-gap asymmetry highlighted in Section~\ref{sec:introduction} appears already on the raw Claude Code variant (Dx 4.22, Disp 4.53 vs.\ CA 2.69, Timeliness 2.65) and persists on the baseline framework (Dx 4.39, Disp 4.52 vs.\ CA 2.94, Timeliness 3.34), so the gap is a property of how current agents handle multi-patient execution rather than an artifact of the framework. \OURAGENT{} closes this asymmetry on top of the baseline, raising overall by $+0.30$ ($p<0.01$), critical actions by $+0.73$ ($+25\%$, $p<0.001$), and timeliness by $+0.45$ ($+13\%$, $p<0.01$); diagnosis and disposition stay within $\pm 0.03$ of the baseline ($p > 0.8$), confirming that the gains come from execution rather than from changes in clinical judgment.

These aggregates pool the six search batches, over which the operating manual was evolved, with the four held-out batches. Refitting the model on the held-out batches alone ($n = 48$) leaves critical actions significant ($+0.625$, $+22\%$, $p = 0.024$), with timeliness ($+0.312$) and overall ($+0.193$) positive but not significant (Table~\ref{tab:heldout}); we treat the held-out critical-actions result as our primary quantitative claim. Re-grading under the four validation judges reproduces the same pattern on every judge (Table~\ref{tab:multijudge}), including one that shares a model family with neither the evaluated agent nor the Patient Engine, and three of the four report a larger timeliness gain than the primary judge, so the reported improvement is not specific to GPT-4.1.

\subsection{Component Ablations}
\label{sec:results-ablation}

Figure~\ref{fig:ablation} adds the three components (harness, skills, acting subagents) one at a time on top of the baseline framework (mixed-effects differences and BH-FDR-corrected significance markers in Table~\ref{tab:significance}, Appendix~\ref{app:results-significance}). Each component yields a positive mean shift, but with different magnitudes. Pooled across all ten batches, the self-evolving \textbf{harness} shows the largest single-component gain ($+0.18$ overall, $+0.38$ critical actions, $+0.34$ timeliness), and is the only single-component variant whose gains on critical actions and timeliness clear the BH-FDR-corrected significance threshold. \textbf{Skills} contributes the second-largest gain ($+0.12$ overall, $+0.22$ on critical actions and timeliness), and \textbf{subagents alone} the smallest ($+0.04$ overall); neither clears the significance threshold on any individual dimension. This ordering reverses on the held-out batches, where $+H$ reaches 3.82 against a baseline of 3.78 and $+SA$ becomes the strongest single component at 3.97 (Table~\ref{tab:search-heldout}).

The three components compose: the full stack reaches $4.10$ overall, with critical actions and timeliness exceeding any single-component variant by $\geq 0.10$ points, indicating the components are not substitutable for one another. Holding skills and subagents fixed and varying only the operating manual, the evolved manual adds $+0.12$ overall over S+SA on all batches ($p = 0.36$) and $+0.02$ on held-out ($p = 0.88$; Appendix~\ref{app:marginal-harness}), so its marginal contribution on top of the architectural components is not statistically separable. Diagnosis and disposition remain within $\pm 0.11$ of the baseline across all variants and never reach significance, consistent with the components being designed to repair execution rather than knowledge.

\begin{figure*}
    \centering
    \includegraphics[width=1\linewidth]{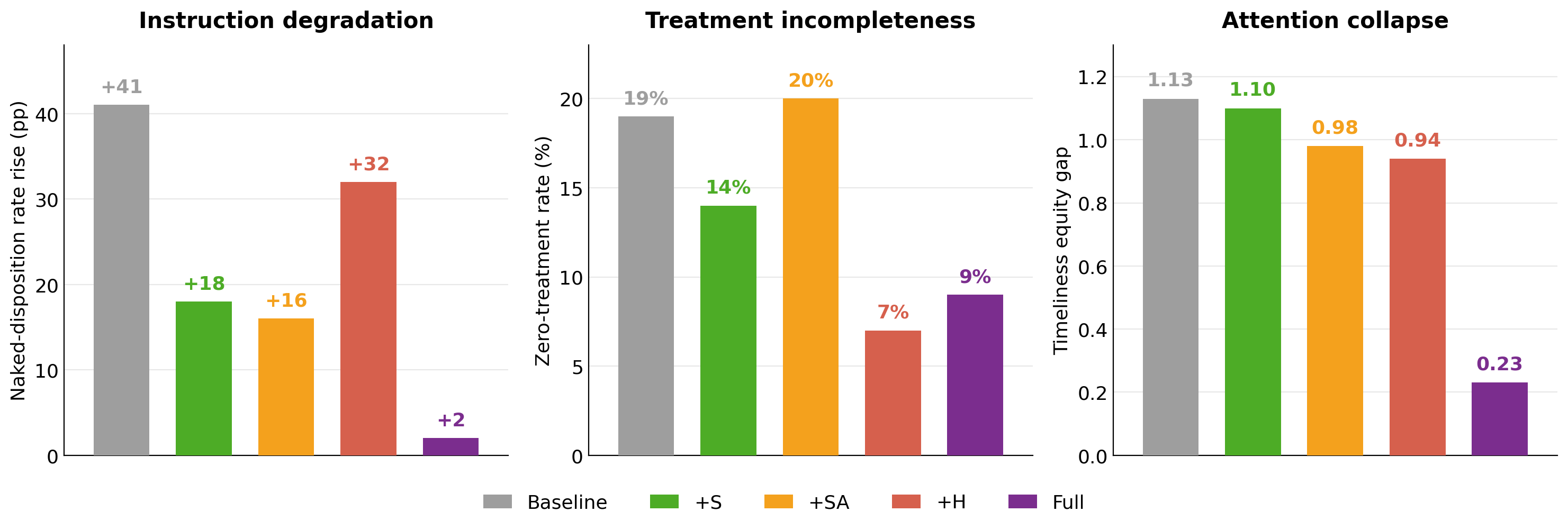}
    \caption{\textbf{Per-component reduction of targeted failure modes.} Three panels: (left) naked-disposition rate early (hours 0--2) versus late (hours 4--6); (middle) zero-treatment rate among correctly-diagnosed patients; (right) timeliness gap between non-severe and severe patients. Each compares the baseline, single-component variants ($+S$, $+SA$, $+H$), and full \OURAGENT{}. Lower is better.}
    \label{fig:failure-modes}
\end{figure*}

\subsection{Failure-Mode Analysis}
\label{sec:results-failure-modes}

Figure~\ref{fig:failure-modes} reports the three per-trace failure-mode counters introduced in Section~\ref{sec:introduction}, giving a per-trace view of \emph{where} each component intervenes and complementing the aggregate scores; Table~\ref{tab:failure-modes} in Appendix~\ref{app:results-failure-modes} lists the exact per-configuration values. Each failure mode is operationalized with one concrete per-trace counter. Instruction adherence can lapse in many ways and admits no single objective measure, so \emph{Drift} tracks one frequent, unambiguous violation: the \emph{naked disposition}, a patient dispositioned with $\leq 1$ prior treatment order, in violation of the manual's treatment-first rule. We report its rate early (hours 0--2) versus late (hours 4--6), so the rise captures decay over the shift. The comparison is within-subject: both configurations see the same twelve patients in the same arrival order on every batch, so any difference in case mix or queue depth between early and late hours applies equally to both, and a rise in one system but not the other cannot be attributed to shifting task difficulty. \emph{Incomplete} is the zero-treatment rate: the fraction of correctly-diagnosed patients given no disease-specific treatment. \emph{Neglect} is the timeliness equity gap: the mean-timeliness difference between non-severe and severe patients, where severe is defined by the judge's holistic encounter rating on the baseline trace ($\leq 2$, $n = 20$; Appendix~\ref{app:results-failure-modes}).

Each component reduces the counter for the failure mode it targets: the harness slows instruction-adherence drift, cutting the early$\to$late rise in naked dispositions from $+41$pp to $+32$pp; the skills library lowers the zero-treatment rate from 19\% to 14\%; and the subagents narrow the timeliness equity gap from 1.13 to 0.98. The effects do not decompose cleanly across components, however: empirically, the skills library and the subagents reduce drift even further than the harness does (to $+18$ and $+16$pp, versus $+32$pp), and the harness in turn attains the lowest zero-treatment rate of any single component (7\%). This pattern is consistent with the three failure modes acting as a \emph{coupled bottleneck}: each component spills onto multiple counters, and no single component closes the execution gap on its own.

The decisive reductions on all three counters appear only when the components are combined. Full \OURAGENT{} nearly eliminates the drift (a $+2$pp early$\to$late rise, far below any single-component variant; 16\%$\to$57\% for the baseline versus 10\%$\to$12\% for \OURAGENT{} on the same patients), collapses the timeliness equity gap from 1.13 to 0.23, and roughly halves the zero-treatment rate (19\% to 9\%). Taken together, the per-trace counters corroborate the aggregate ablation: the three components are individually partial and jointly sufficient, and the coupled bottleneck only releases when the self-evolving harness, the skills library, and the isolated subagents act together.

\begin{figure}
    \centering
    \includegraphics[width=1\linewidth]{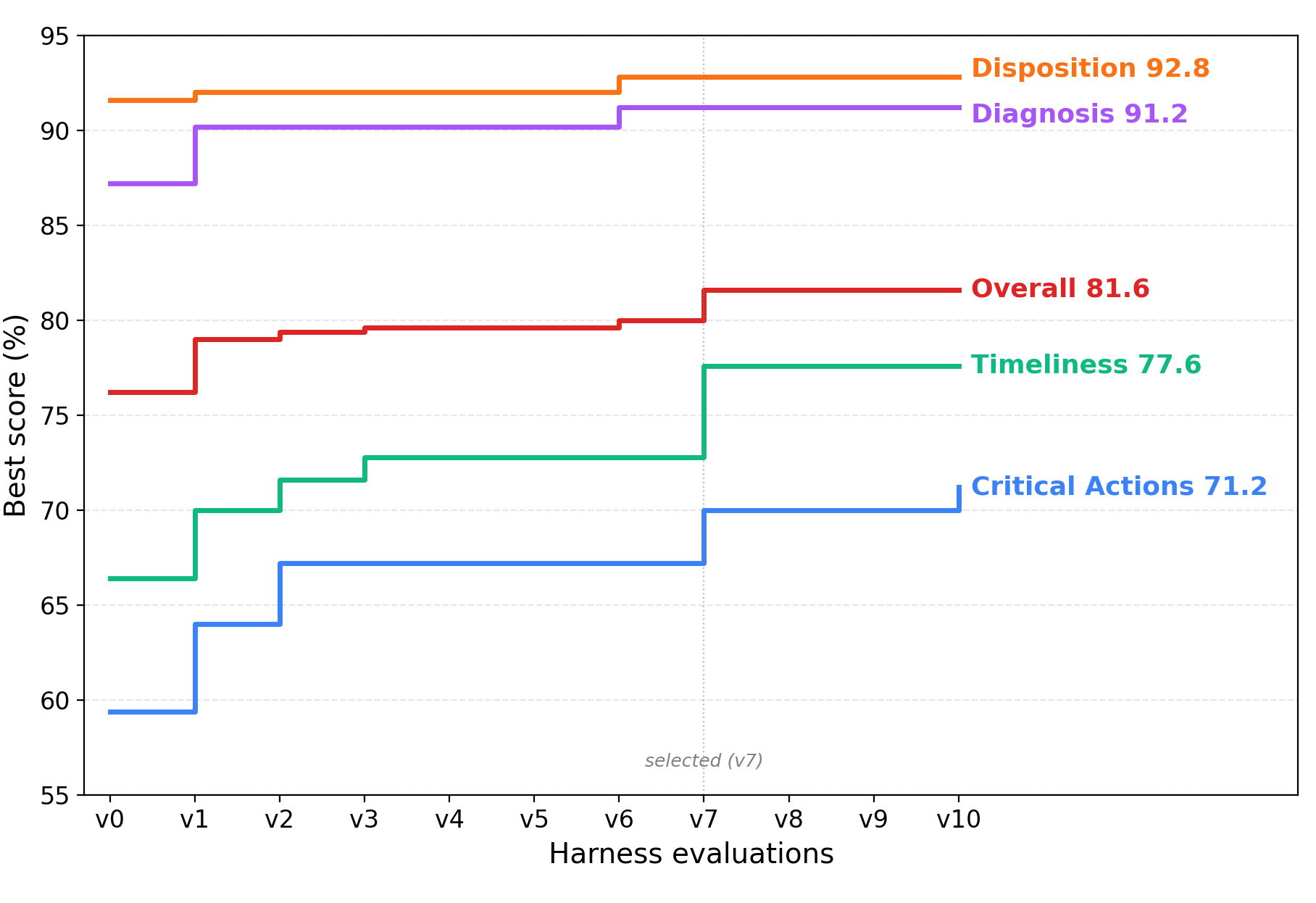}
    \caption{\textbf{Self-evolving harness search trajectory.} Best-so-far mean scores on the six search batches across 10 proposer iterations, one line per CES dimension. v7 is selected on overall search-batch score and used in \OURAGENT{}; v6--v8 form a high-performing plateau.}
    \label{fig:harness-trajectory}
\end{figure}

\subsection{Harness Search Trajectory}
\label{sec:results-harness}

Figure~\ref{fig:harness-trajectory} shows the best-so-far search-batch scores across the 10 proposer iterations, and Table~\ref{tab:harness-trajectory} in Appendix~\ref{app:results-harness-trajectory} reports the raw per-iteration numbers. The trajectory is non-monotone: critical actions and timeliness rise quickly from v0 to v2, plateau through v4--v5, and reach a high-performing plateau at v6--v8. We select v7 as the final manual on the basis of overall search-batch score. Iterations v9--v10 regress on diagnosis or disposition. Inspecting the diff from v0 to v7 reveals two classes of edit: (i)~a procedural edit adding a pre-disposition treatment-enumeration routine tied to the seven treatment categories, and (ii)~knowledge edits to the clinical content of the manual, with 13 new presentation rows (e.g., pulmonary embolism) and roughly 7 enriched management entries, such as correcting prophylactic to therapeutic anticoagulation dosing. These additions are presentation-level decision rules, not scenario-specific answers. To reduce overfitting to the search batches, the proposer is instructed to add only guidance that generalizes across cases and to act only on failure patterns recurring across multiple patients and batches. The full configuration retains a $+0.19$ overall gain over the baseline on the held-out batches the proposer never observes (Table~\ref{tab:search-heldout}).

At inference time, full \OURAGENT{} requires $1.21\times$ the LLM calls of the baseline (433 vs.\ 359 per batch) and $1.52\times$ the wall-clock (118 vs.\ 78 minutes per batch), so the gains do not come from a large increase in inference compute. Harness evolution is a separate, one-time development cost of 10 iterations, each a proposer step ($\approx$10 minutes) followed by an evaluation rollout over the six search batches, producing a frozen manual that is reused unchanged at inference.

\begin{table}[t]
\centering
\caption{\textbf{Subagent advisory vs.\ acting variants.} Mean scores under the full \OURAGENT{} configuration. The acting variant gives each subagent direct MCP access within its scope; the advisory variant returns structured recommendations parsed by the Main Agent. We use the acting variant as the default.}
\label{tab:subagent-variants}
\small
\setlength{\tabcolsep}{4pt}
\begin{tabular}{lccccc}
\toprule
Variant & Overall & Dx & CA & Time & Disp \\
\midrule
Advisory          & 4.00 & \textbf{4.39} & 3.41 & 3.69 & 4.49 \\
Acting (default)  & \textbf{4.10} & 4.38 & \textbf{3.67} & \textbf{3.79} & \textbf{4.54} \\
\bottomrule
\end{tabular}
\end{table}

\subsection{Subagent Variants}
\label{sec:results-subagent}

Table~\ref{tab:subagent-variants} compares the advisory and acting variants of the subagent decomposition (Section~\ref{sec:method-subagents}) under the full \OURAGENT{} configuration. The acting variant outperforms the advisory variant by +0.10 overall, +0.26 on critical actions, and +0.10 on timeliness, with no degradation on diagnosis or disposition. The gap is driven by tool-call efficiency: when subagents place their own orders, the Main Agent avoids paraphrasing and re-resolving long structured recommendations, removing a class of vocabulary-resolution failures observed in the advisory traces. We use the acting variant as the default elsewhere in the paper.

\subsection{Qualitative Analysis}
\label{sec:results-qualitative}

We walk through a representative \OURAGENT{} rollout in which the inner-loop subagents intercept a failure that the baseline framework commits on the same patient. Patient B1-P10 presents in atrial flutter with rapid ventricular response (RVR); both systems reach the correct diagnosis. On the baseline trace, the Main Agent orders heparin and fluids but fails to initiate a rate-control strategy or a durable oral anticoagulation plan before disposition. The patient remains in rapid atrial flutter and worsens over subsequent turns, with escalating tachycardia and hypoxemia (heart rate 135$\to$153, SpO$_2$ 95\%$\to$86\%; critical-actions score 1/5), consistent with persistent RVR in the setting of cardiopulmonary vulnerability and empiric fluid administration. On the \OURAGENT{} trace, the treatment-completion subagent flags the missing rate-control strategy and post-ED anticoagulation plan. The Main Agent adds diltiazem, after excluding hypotension, pre-excitation, and decompensated systolic heart failure, and starts apixaban for indicated stroke prevention. The patient's ventricular rate improves, oxygenation stabilizes, and the patient is admitted to cardiology with a complete regimen (critical-actions score 1$\to$5). More traces with analogous interceptions are reported in Appendix~\ref{app:qualitative-traces}.
\section{Conclusion}
\label{sec:conclusion}

We studied the execution gap in long-horizon, multi-task agentic settings, using the Clinical Environment Simulator as a testbed in which the phenomenon is measurable: current agents reach the correct diagnosis in most cases yet fail to deliver complete and timely critical actions under sustained shift-long load. We operationalized this gap with three per-trace counters that target long-horizon failure modes (instruction-adherence drift, treatment incompleteness, severity-equity gap), and instantiated an adaptive agent scaffolding, Asclepius, with a component for each: a self-evolving harness that rewrites the operating manual between shifts, an externalized clinical skills library, and three isolated subagents that partition per-turn decisions across the patient queue. On held-out batches never observed during harness evolution, \OURAGENT{} improves critical-action correctness by 22\% over a strong baseline framework
while preserving diagnostic accuracy, with consistent gains across five LLM
judges from three model families; on the full ten-batch set, improvements
reach 25\% on critical actions and 13\% on timeliness. The strongest empirical finding is structural: the three failure modes form a coupled bottleneck, in that no single component closes the execution gap on its own, and the decisive reductions on all three counters appear only when the self-evolving harness, the skills library, and the isolated subagents act together. For clinical AI evaluation, the result argues for benchmarks that measure sustained, complete, and equitable execution across a patient queue, not diagnostic accuracy on isolated cases.

\section{Limitations}
\label{sec:limitations}

\paragraph{LLM-based grading.} All four CES dimensions are scored by LLM
judges. Our primary judge (GPT-4.1; Appendix~\ref{app:evaluation}) shares a
model family with the simulator's Patient Engine, so correlated errors between
generated physiology and its grading are a structural concern. Re-grading under
four further judges across three model families reproduces the gains
(Table~\ref{tab:multijudge}), indicating the improvement is not specific to one
grader. Absolute scores nonetheless inherit any miscalibration or ceiling
effects shared across LLM graders, and we did not run an expert-agreement study
on \OURAGENT{} traces. Gains should be read as improvements under automated
grading, not as validated clinical outcomes.

\paragraph{Single rollout per configuration and batch.} Each configuration is
run once on each batch, so every reported cell rests on one trajectory of a
stochastic agent in a stochastic simulator. Our mixed-effects model treats
patient and batch as random effects and therefore quantifies patient- and
batch-level heterogeneity, not run-to-run execution variance. Two observations
bound how much this drives the results: full \OURAGENT{} improves critical
actions on all ten batches individually (smallest margin $+0.17$;
Appendix~\ref{app:results-per-batch}), and the same direction and approximate magnitude
are reproduced by four independent judges. Neither substitutes for repeated
execution, and per-batch overall differences range from $-0.12$ to $+0.71$, so
individual per-batch values and the ranking within the v6--v8 plateau should
not be over-read.

\paragraph{Simulation-to-reality gap.} CES is a simulator: patient cards are
derived from de-identified records and physiology is LLM-generated. Performance
on CES is a proxy for, not evidence of, real clinical competence. We evaluate a
single default ED configuration (four beds, six nurses, one physician, six-hour
shift); robustness to other shift lengths, staffing ratios, and acuity mixes is
untested, as is generalization beyond the English-language, guideline-dosing
regime encoded in the skills library. All results use a single backbone
(Claude Opus 4.6) in a single environment; whether the components help weaker
base models, or transfer to other long-horizon multi-objective environments,
is untested.

\paragraph{Cost and scalability of harness evolution.} The self-evolving outer
loop requires repeated multi-hour rollouts across six search batches per
iteration plus a separate proposer agent, making it computationally expensive
relative to a hand-written manual. Selection uses only six search batches;
although the held-out batches confirm generalization, the overall gain
compresses from $+0.36$ on search to $+0.19$ on held-out, indicating some
adaptation to the search trajectories. The skills library is hand-curated rather
than learned, so its coverage is bounded by manual effort and may omit rare
presentations.

\bibliography{reference}

\clearpage
\appendix

\section{CES Platform Details}
\label{app:simulator}

This appendix expands on the brief description of the Clinical Environment Simulator (CES)~\cite{luo2026clinical} in Section~\ref{sec:exp-ces} and supplies the simulator mechanics that our results in Section~\ref{sec:results} depend on.

\subsection{Temporal Model and Shift Structure}
\label{app:simulator-time}

CES is a turn-based discrete-event simulator in which one turn corresponds to five simulated minutes. A default scenario spans a six-hour emergency-department shift, that is, 72 turns. At each turn the engine executes a fixed-order pipeline: check for scheduled patient arrivals, auto-assign beds from the queue, process all pending actions, and update patient vital signs. Clinical actions submitted during a turn are queued as pending actions and resolved in batch at the corresponding turn boundary, ensuring deterministic ordering and preventing within-turn causal inconsistencies.

A scenario comprises twelve patients with staggered arrival times across the shift, simulating realistic emergency-department patient flow. The staggered arrivals force the agent to manage time across concurrent encounters: excessive time spent on a single patient causes other patients to deteriorate past their hidden thresholds. Patient cards are derived from de-identified emergency-department visit records and converted into structured YAML scenarios containing demographics, a triage-categorized presentation, a patient summary with separated subjective history and objective examination findings, clinical background, initial vital signs, a ground-truth diagnosis, the list of required critical actions, the acceptable disposition options, and a hidden deterioration threshold.

\subsection{Patient Severity and Deterioration}
\label{app:simulator-severity}

Each patient is assigned one of three severity classes that govern temporal dynamics: \emph{mundane} patients carry no physiological risk and do not deteriorate; \emph{dynamic moderate} patients (e.g., sepsis without shock, community pneumonia) face gradual deterioration risk with a hidden threshold randomly sampled in 90--120 minutes; \emph{dynamic severe} patients (e.g., hemorrhagic shock, intracranial hemorrhage) face rapid deterioration risk with a hidden threshold in 30--60 minutes. Prior to the threshold, vital signs oscillate around the patient's initial set-point with severity-scaled corridors and a mean-reversion factor; after the threshold elapses without the required critical actions, pathological drift begins through three severity phases (early, moderate, severe). When the required critical actions are completed, drift ceases and vital signs recover along pharmacokinetic time courses appropriate to the administered interventions. Only dynamic severe patients can trigger a code-blue event when vitals breach boundary values; code blue consumes 20--30 minutes of physician time and forces immediate transfer to the intensive care unit with reduced encounter scores.

\subsection{Action Costs and Resource Constraints}
\label{app:simulator-actions}

Action durations are calibrated against expert physician consensus. A comprehensive history-and-physical examination consumes 10 minutes (2 turns) of physician time; a follow-up or targeted H\&P requires 5 minutes (1 turn). Triage assessment is auto-generated on arrival. Medication administration requires 5 minutes (1 turn) of nurse time for standard medications; transfusions and large-volume infusions require 20 minutes (4 turns). Diagnostic test latencies vary by modality: laboratory studies require 15--125 minutes (e.g., basic metabolic panel 30~min, coagulation studies 45~min, PCR panels 125~min), with culture results requiring approximately 24 hours; imaging studies require 5--75 minutes depending on modality (bedside ultrasound 5~min, plain radiography 10--15~min, magnetic resonance imaging 40~min). Procedural durations vary similarly (arterial line 15~min; chest tube 20~min; central venous catheter 25~min).

The default emergency-department configuration comprises 4 beds, 6 nurses, 1 physician (the agent under evaluation), 14 specialist consultants spanning 9 medical and 5 surgical specialties, one unit each of CT/MRI/X-ray/ultrasound/ECG, and 1 laboratory analyzer. Beds, staff, imaging equipment, and laboratory throughput are managed as capacitated queues with priority-based allocation by clinical acuity (HIGH before MEDIUM before LOW). Certain actions (e.g., oxygen administration, intravenous medications) require a bed assignment, and the finite bed supply creates realistic bottlenecks. Specialist consults have stochastic response latency (2--4 turns) and consultation duration (2--12 turns); consultants receive only the information available to the agent at the time of request and never the ground-truth diagnosis.

\subsection{Medication and Treatment Effects}
\label{app:simulator-medication}

To prevent unrealistic vital-sign discontinuities, treatment effects are categorized by onset profile. Immediate-onset interventions (5--10~min) include intravenous fluid resuscitation and supplemental oxygen. Delayed-effect medications (15--30~min) include antibiotics (halting physiological deterioration) and bronchodilators. Gradual-effect agents ($\geq 30$~min) include antipyretics. This temporal fidelity penalizes delayed treatment initiation and rewards early, appropriate intervention.

\subsection{Engine Architecture}
\label{app:simulator-engine}

CES is organized around two decoupled engines that communicate through a typed message bus. The Patient Engine maintains each patient's clinical state (vital signs, active diagnoses, medication effects, deterioration trajectory) and delegates physiological reasoning to a fixed Patient LLM (GPT-4.1 \citep{openai2025gpt41}) that generates vital-sign updates, history-and-physical narratives, and diagnostic results at each turn. The Hospital Engine manages capacitated resources, tracking allocation, queuing, and release through priority-based scheduling. A turn-based orchestrator coordinates the two engines: at turn start it broadcasts a turn-start signal; the Hospital Engine processes resource allocations and order completions and notifies the Patient Engine of newly available results; the Patient Engine incorporates results, processes arrivals, and advances physiological trajectories; at turn end the orchestrator emits a turn-end signal with all generated events. Agents interact with the simulator through a browser-based interface served by a session-based REST API, with a patient tracking board, individual patient charts, an order catalog of 6,888 items, and real-time event notifications.

\section{Baseline Agent Configuration Details}
\label{app:baseline}

Our baseline configuration wraps the underlying LLM with a CES-customized harness exposing 17 clinically meaningful tools organized in an Observe--Act--Control trichotomy through the Model Context Protocol (MCP). The same harness is the starting point for \OURAGENT{}; the components in Section~\ref{sec:method} are added on top.

\subsection{Tool Categories}
\label{app:baseline-tools}

\paragraph{Observation tools (5).} Return structured representations of the simulator state to the agent rather than requiring it to interpret raw rendered content: the patient tracking board, individual patient summaries (chief complaint, vitals, key findings), order histories, complete patient records, and event logs. The agent receives organized vital signs, pending orders, and diagnostic results directly, mirroring the information a physician extracts at a glance from an electronic-health-record display.

\paragraph{Action tools (8).} Encapsulate multi-step clinical workflows as single tool calls: order tests, order medications, order procedures, order consultation, discharge, admit, transfer, and perform history-and-physical examination. Each action tool accepts natural clinical language (e.g., ``CBC'', ``morphine'', ``CT head'') and resolves it to exact catalog entries through the vocabulary resolution pipeline described below.

\paragraph{Control tools (4).} Manage simulator lifecycle: session initialization (automatic login and scenario loading), single-step turn advancement, run-until-event turn advancement, and blocking-state handling for code blue, rapid-response-team, and physician-busy states.

\subsection{Medical Vocabulary Resolution}
\label{app:baseline-vocab}

Free-text order strings are resolved to catalog entries through a three-stage pipeline. \emph{Stage 1} performs deterministic lookup against 464 hand-curated abbreviation mappings (e.g., ``CBC'' $\to$ ``complete blood count''). \emph{Stage 2}, invoked when Stage 1 fails, embeds the query with a sentence-transformer model and searches a vector database containing 5,823 medications, 943 diagnostic tests, and 122 procedures, accepting the best semantic match above a cosine-similarity threshold of 0.3. \emph{Stage 3} performs interface-level verification: the resolved name is entered into the simulator's search interface, and fuzzy string matching (weighted-ratio scorer, cutoff 60) is applied against the displayed options to confirm that the ordered item exists in the current simulator state. This pipeline handles the vocabulary mismatch between clinical shorthand and the simulator's catalog of 6,888 orderable items.

\subsection{Pre-Action Guards and the Operating Manual}
\label{app:baseline-manual}

Tool descriptions and pre-action guards encode workflow knowledge directly into the interface: patients must be assigned a bed before medications can be ordered, intravenous formulations are recommended over oral for acute presentations, triage is enforced by acuity (HIGH before MEDIUM before LOW), and disposition tools automatically surface the current order set so the agent can audit care delivered before submission. The operating manual instructs the agent on triage, parallel workup, treatment, and disposition, including completeness rules (``treatment-first principle'', ``no naked diagnosis'', a seven-category treatment-completeness sweep, and a pre-disposition verification checklist). In our experiments the manual is the starting point $M_0$ from which the self-evolving harness (Section~\ref{sec:method-harness}) iterates.

\section{Asclepius Component Artifacts}
\label{app:artifacts}

The complete prompts, skill modules, and subagent definitions for every configuration we evaluate are released at \url{https://github.com/rajpurkarlab/Asclepius}. The repository is
organized by configuration, mirroring the ablations in
Section~\ref{sec:exp-configs}: the baseline framework, each single-component
variant, and full \OURAGENT{} in both acting and advisory forms. Each
configuration directory contains the main agent prompt (the hand-crafted
operating manual $M_0$ for the baseline, or the evolved manual $M^*$ for the
harness and full configurations); where applicable, the skills library, the acting or advisory subagent definitions (triage prioritizer,
diagnostician, and treatment checker), and the proposer prompt that evolves the
operating manual (Section~\ref{sec:method-harness}).

\section{Evaluation Design}
\label{app:evaluation}

Each shift is graded by a fixed LLM judge (GPT-4.1 \citep{openai2025gpt41}) configured with structured prompts, with the four-dimension rubric described below~\cite{luo2026clinical}. The judge receives a ground-truth answer key derived from the patient card (ground-truth diagnosis, severity class, required critical actions, tests required for diagnosis, acceptable disposition options, hidden deterioration threshold) together with a comprehensive record of the agent's actions (submitted diagnosis, medications administered with timestamps, procedures performed, diagnostic tests ordered, disposition decision and discharge instructions, and any code-blue or rapid-response-team activations). For clinical context the judge also receives the patient's initial presentation, the full vital-sign trajectory, a chronological log of clinical events, returned test results, and any specialist consultations requested. The judge never receives the Patient Engine's internal prompts. Performance is scored on a 1--5 scale along four dimensions.

\paragraph{Diagnosis.} Evaluated under a clinical-equivalence matching framework. A score of 5 denotes an exact or clinically equivalent match to the ground truth; 4 reflects a clinically defensible alternative diagnosis (e.g., pericarditis for a ground truth of myocarditis); 3 indicates correct organ-system identification without sufficient specificity, or a multi-part diagnosis containing incorrect extra information; 2 corresponds to an incorrect but non-dangerous diagnosis; 1 is reserved for cases in which a life-threatening diagnosis is missed and the proposed management would cause harm. A non-specific diagnosis when specificity is clinically required is a scoring failure (e.g., ``chest pain'' for ST-elevation myocardial infarction).

\paragraph{Critical Actions.} Scored on the fraction of required actions completed. The judge applies clinical-equivalence matching for synonymous terminology (e.g., acetaminophen $\equiv$ paracetamol; normal saline bolus $\equiv$ 0.9\% NaCl infusion). Medication dosing is evaluated against standard-of-care ranges rather than requiring exact values.

\paragraph{Timeliness.} Assesses whether critical actions are completed before the hidden deterioration threshold. If required critical actions remain incomplete, the maximum achievable timeliness score is capped at 3 regardless of speed. Activation of a rapid response team with a correct working diagnosis for a deteriorating patient may earn a 2; a code-blue event in the absence of any prior critical actions receives a 1.

\paragraph{Disposition.} Assesses whether the final placement (discharge, admission to a specific service, or transfer, e.g., to the operating room or catheterization laboratory) matches the acceptable disposition options for the case.

For every encounter we report an \emph{overall} score, defined as the unweighted mean of the four per-dimension scores above; this is the overall score used in every table and figure. The LLM judge additionally produces a single \emph{holistic encounter rating}. We use it for identifying the severe subset in the failure-mode analysis (Appendix~\ref{app:results-failure-modes}). Encounters are scored independently per patient and then aggregated to per-batch and per-configuration means; we use these aggregates throughout Section~\ref{sec:results}.
\subsection{Judge Reliability}
\label{app:evaluation-judge-reliability}
The CES judge carries prior validation reported by \citet{luo2026clinical}. On the simulator's plausibility checks, an internal-medicine physician blinded
to the automated results, patient cards, and Patient Engine prompts reviewed
2{,}802 matched items across 78 scenarios, agreeing with the automated checker
at 92.5\% raw agreement (Gwet's AC1 = 0.92, 95\% CI 0.91--0.93). On the
four-dimension scoring rubric we use, a blinded rater scored 48 encounters
drawn from both physician and agent sessions, yielding Gwet's AC2 of 0.93
(diagnosis), 0.86 (critical actions), and 0.86 (timeliness), with an overall
AC2 of 0.88. This indicates that the judge's scores on the dimensions central
to our claims track expert judgment closely, and that the checker carries a
conservative bias toward flagging implausible outputs rather than rewarding
them.

Two points remain specific to our setting. First, these agreement studies
validate the judge on CES trajectories broadly, not on the \OURAGENT{}
configurations evaluated here; we did not run an expert-agreement study on our
own traces. Second, the primary judge and the simulator's Patient Engine are
the same model family (GPT-4.1). We address this directly in
Section~\ref{sec:results-main} by re-grading all 120 patients under the
baseline and full configurations with four further judges spanning three model
families; all five reproduce the reported gains
(Table~\ref{tab:multijudge}).

\subsection{Patient Batches}
\label{app:evaluation-batches}

Each patient batch comprises twelve patients with varying acuities, arrival times, and underlying diagnoses, sampled to span the acuity and pathophysiology spectrum. We use ten batches in total. Batches 1--6 are \emph{search batches}, used by the self-evolving harness as the substrate over which the proposer iterates. Batches 7--10 are \emph{held-out validation batches}, never seen by the proposer at any iteration. Each configuration is run once per batch, with per-dimension scores averaged across the batches in each split.

\section{Per-Batch, Search-Split, and Significance Results}
\label{app:results-per-batch}

This appendix expands the aggregate results in Section~\ref{sec:results} by reporting (i) the search-batch (1--6) and held-out (7--10) subsets, (ii) the full per-batch overall scores, and (iii) the full BH-FDR-corrected significance table from the mixed-effects model. 

\subsection{Search-Batch and Held-Out Means}
\label{app:results-search-split}


Table~\ref{tab:search-heldout} reports the means for the five real configurations on the search batches (1--6) and on the held-out batches (7--10) separately. Full \OURAGENT{} is the best configuration on both splits, with held-out overall $3.97$ vs.\ baseline $3.78$ ($\Delta = +0.19$). The overall gain compresses from $+0.36$ on search to $+0.19$ on held-out, a roughly 47\% compression. We read this compression as informative rather than as a failure of generalization: the proposer adapts in part to recurring patterns of the search batches, and the persistent $+0.19$ on never-observed batches gives a lower bound on the component of the gain that is generalizable. Within the single-component variants, the harness ranks first on the search subset but loses that lead on the held-out subset (where $+SA$ ranks highest among single components), consistent with the same picture: the self-evolving harness carries some search-set-specific signal that does not fully transfer, while the architectural components ($+SA$, skills) carry gains that transfer more uniformly.

\begin{table*}[h]
\centering
\caption{\textbf{Search-batch and held-out means.} Mean overall and per-dimension scores on batches 1--6 (search) and 7--10 (held-out) for the five ablation configurations. Overall is the unweighted mean of the four sub-dimension scores. Bold marks the best score in each column.}
\label{tab:search-heldout}
\small
\setlength{\tabcolsep}{4pt}
\begin{tabular}{l cc cc cc cc cc}
\toprule
& \multicolumn{2}{c}{Overall} & \multicolumn{2}{c}{Diagnosis} & \multicolumn{2}{c}{Critical Actions} & \multicolumn{2}{c}{Timeliness} & \multicolumn{2}{c}{Disposition} \\
Configuration & 1--6 & 7--10 & 1--6 & 7--10 & 1--6 & 7--10 & 1--6 & 7--10 & 1--6 & 7--10 \\
\midrule
Baseline       & 3.81 & 3.78 & 4.36 & 4.44 & 2.97 & 2.90 & 3.32 & 3.38 & 4.58 & 4.42 \\
\, + $S$       & 3.93 & 3.91 & 4.50 & 4.42 & 3.18 & 3.13 & 3.54 & 3.58 & 4.51 & 4.50 \\
\, + $SA$      & 3.76 & 3.97 & 4.39 & 4.50 & 2.94 & 3.35 & 3.28 & 3.63 & 4.42 & 4.40 \\
\, + $H$       & 4.08 & 3.82 & 4.51 & 4.38 & 3.50 & 3.04 & 3.86 & 3.42 & 4.43 & 4.46 \\
Full \OURAGENT{}          & \textbf{4.17} & \textbf{3.97} & 4.39 & 4.33 & \textbf{3.76} & \textbf{3.54} & \textbf{3.86} & \textbf{3.69} & \textbf{4.68} & 4.33 \\
\bottomrule
\end{tabular}
\end{table*}

\subsection{Per-Batch Overall Scores}
\label{app:results-per-batch-overall}

Table~\ref{tab:per-batch} reports the overall score for the baseline framework and full \OURAGENT{} on each of the ten batches individually, so per-batch variance can be inspected. Full \OURAGENT{} beats the baseline on overall in 8 of 10 batches; the two exceptions, B3 and B6, are small ($-0.12$ and $-0.06$). On critical actions, full \OURAGENT{} beats the baseline on every batch (smallest per-batch margin $+0.17$).

\begin{table*}[h!]
\centering
\caption{\textbf{Per-batch overall scores.} Overall (mean across the four dimensions) for the baseline framework and full \OURAGENT{} on each of the ten batches.}
\label{tab:per-batch}
\small
\begin{tabular}{lcccccccccc}
\toprule
Configuration & B1 & B2 & B3 & B4 & B5 & B6 & B7 & B8 & B9 & B10 \\
\midrule
Baseline             & 3.54 & 3.85 & 4.06 & 3.58 & 3.73 & 4.08 & 3.56 & 3.79 & 3.94 & 3.83 \\
Full \OURAGENT{}     & 4.25 & 4.35 & 3.94 & 4.23 & 4.25 & 4.02 & 3.88 & 3.92 & 4.21 & 3.90 \\
\midrule
$\Delta$             & $+0.71$ & $+0.50$ & $-0.12$ & $+0.65$ & $+0.52$ & $-0.06$ & $+0.31$ & $+0.12$ & $+0.27$ & $+0.06$ \\
\bottomrule
\end{tabular}
\end{table*}

\subsection{Component Ablation Means and Significance Tests}
\label{app:results-significance}

Table~\ref{tab:ablation} reports the absolute per-dimension means for each
ablation configuration; Table~\ref{tab:significance} gives the corresponding
mixed-effects mean differences and BH-FDR-corrected $p$-values.

We fit a patient-level mixed-effects model $y \sim \mathrm{system} + (1 \,|\, \mathrm{patient}) + (1 \,|\, \mathrm{batch})$ with the baseline framework as the reference, Satterthwaite degrees of freedom, and BH-FDR correction across all 20 system-by-dimension comparisons. Table~\ref{tab:significance} reports the corrected results. Three patterns stand out. First, full \OURAGENT{} is significant on every execution-related dimension (overall, critical actions, timeliness) at $p<0.01$ or stricter. Second, among single-component variants, the harness is the only one to reach the corrected threshold on any dimension (critical actions and timeliness); skills and subagents alone reach significance on none. Third, diagnosis and disposition do not significantly change for any configuration, consistent with all systems near the judge's ceiling.

\begin{table}[h!]
\centering
\caption{\textbf{Component ablation across all ten patient batches.} Per-dimension mean scores when adding the harness ($H$), skills ($S$), and acting subagents ($SA$) on top of the baseline framework. Significance markers (vs.\ baseline, BH-FDR corrected) follow the overall and execution columns. Bold marks the best score in each column.}
\label{tab:ablation}
\small
\setlength{\tabcolsep}{4pt}
\begin{tabular}{llllll}
\toprule
Configuration & Overall & Dx & CA & Time & Disp \\
\midrule
Baseline                  & 3.80             & 4.39 & 2.94 & 3.34 & 4.52 \\
\, + $S$                  & 3.92\textsuperscript{ns}         & 4.47 & 3.16 & 3.56 & 4.51 \\
\, + $SA$                 & 3.84\textsuperscript{ns} & 4.43 & 3.11 & 3.42 & 4.41 \\
\, + $H$                  & 3.98\textsuperscript{ns}       & 4.46 & 3.32\,* & 3.68\,* & 4.44 \\
Full \OURAGENT{}      & \textbf{4.10}** & 4.38 & \textbf{3.67}\,*** & \textbf{3.79}\,** & \textbf{4.54} \\
\bottomrule
\end{tabular}
\end{table}

\begin{table*}[h!]
\centering
\caption{\textbf{Significance versus baseline (mixed-effects model, BH-FDR corrected).} For each outcome and configuration, we report the estimated mean difference vs.\ the baseline framework on the 1--5 scale and the BH-FDR-corrected significance (*: $p<0.05$, **: $p<0.01$, ***: $p<0.001$, ns: not significant). Overall is the unweighted mean of the four sub-dimension scores.}
\label{tab:significance}
\small
\setlength{\tabcolsep}{5pt}
\begin{tabular}{l cc cc cc cc}
\toprule
& \multicolumn{2}{c}{+ Skills} & \multicolumn{2}{c}{+ Subagents} & \multicolumn{2}{c}{+ Harness} & \multicolumn{2}{c}{Full} \\
Outcome & $\Delta$ & sig. & $\Delta$ & sig. & $\Delta$ & sig. & $\Delta$ & sig. \\
\midrule
Overall            & $+0.12$ & \textsuperscript{ns} & $+0.04$ & \textsuperscript{ns} & $+0.18$ & \textsuperscript{ns} & $+0.30$ & ** \\
Diagnosis          & $+0.08$ & \textsuperscript{ns} & $+0.04$ & \textsuperscript{ns} & $+0.07$ & \textsuperscript{ns} & $-0.02$ & \textsuperscript{ns} \\
Critical Actions   & $+0.22$ & \textsuperscript{ns} & $+0.17$ & \textsuperscript{ns} & $+0.38$ & *  & $+0.73$ & *** \\
Timeliness         & $+0.22$ & \textsuperscript{ns} & $+0.08$ & \textsuperscript{ns} & $+0.34$ & *  & $+0.45$ & ** \\
Disposition        & $-0.01$ & \textsuperscript{ns} & $-0.11$ & \textsuperscript{ns} & $-0.08$ & \textsuperscript{ns} & $+0.03$ & \textsuperscript{ns} \\
\bottomrule
\end{tabular}
\end{table*}

\subsection{Marginal Contribution of the Evolved Manual}
\label{app:marginal-harness}

The ablation configurations in Figure~\ref{fig:ablation} vary both the manual
version and the set of architectural components. To isolate the manual, we
compare S+SA (skills and acting subagents on $M_0$) against full \OURAGENT{}
(the same components on $M^*$); this contrast holds skills and subagents fixed
and varies only the operating manual. Table~\ref{tab:marginalharness} reports
the mixed-effects mean differences. On all batches the evolved manual adds
$+0.12$ overall, $+0.23$ on critical actions, and $+0.08$ on timeliness; on the
held-out batches the overall difference falls to $+0.02$. Neither split is
statistically separable from zero, so we scope our claims to the system level
rather than attributing an independently significant effect to the harness.

\begin{table}[h]
\centering
\caption{\textbf{Marginal contribution of the evolved manual.} Mixed-effects
mean differences of full \OURAGENT{} ($M^*$) versus S+SA ($M_0$), holding
skills and acting subagents fixed.}
\label{tab:marginalharness}
\small
\begin{tabular}{lcc}
\toprule
Split & $\Delta$ Overall & $p$ \\
\midrule
All batches      & $+0.12$ & 0.364 \\
Held-out (7--10) & $+0.02$ & 0.883 \\
\bottomrule
\end{tabular}
\end{table}

\subsection{Per-Trace Failure-Mode Counters}
\label{app:results-failure-modes}
Table~\ref{tab:failure-modes} gives the per-configuration values of the three
per-trace failure-mode counters visualized in Figure~\ref{fig:failure-modes};
see Section~\ref{sec:results-failure-modes} for the analysis.

\begin{table}[h!]
\centering
\caption{\textbf{Per-trace failure-mode counters across all ten patient batches.} Lower is better for all three counters. \emph{Drift}: decay in the agent's adherence to the operating manual over the shift, measured as the early (hours 0--2)$\to$late (hours 4--6) rise in \emph{naked dispositions} (dispositions placed with $\leq 1$ prior treatment order, violating the manual's treatment-first rule). \emph{Incomplete}: fraction of correctly-diagnosed patients given no disease-specific treatment at all. \emph{Neglect}: timeliness equity gap, the mean-timeliness difference between non-severe and severe patients (severe $=$ baseline holistic encounter rating $\leq 2$, $n{=}20$).}
\label{tab:failure-modes}
\small
\setlength{\tabcolsep}{4pt}
\begin{tabular}{lccc}
\toprule
Config. & Drift (early$\to$late) & Incomplete & Neglect \\
\midrule
Baseline   & 16$\to$57\%            & 19\%           & 1.13 \\
\,+\,$S$   & 10$\to$28\%            & 14\%           & 1.10 \\
\,+\,$SA$  & 13$\to$29\%            & 20\%           & 0.98 \\
\,+\,$H$   & \phantom{0}7$\to$39\%  & \textbf{7\%}   & 0.94 \\
Full       & \textbf{10$\to$12\%}   & \phantom{0}9\% & \textbf{0.23} \\
\bottomrule
\end{tabular}
\end{table}

\subsection{Harness Search Trajectory (Per-Iteration Scores)}
\label{app:results-harness-trajectory}
Table~\ref{tab:harness-trajectory} gives the per-iteration search-batch scores
visualized in Figure~\ref{fig:harness-trajectory};
see Section~\ref{sec:results-harness} for the analysis.
\begin{table*}[h!]
\centering
\caption{\textbf{Harness search trajectory on search batches (1--6).}
Mean scores across the six search batches (out of 5) for each harness
iteration. Overall is the unweighted mean of the four sub-dimension scores. v7 is selected as the final manual based on overall
search-batch performance. Bold marks the best score in each column.}
\label{tab:harness-trajectory}
\small
\begin{tabular}{lccccc}
\toprule
Iteration & Overall & Diagnosis & Critical Actions & Timeliness & Disposition \\
\midrule
v0 (baseline) & 3.81 & 4.36 & 2.97 & 3.32 & 4.58 \\
v1            & 3.95 & 4.51 & 3.20 & 3.50 & 4.60 \\
v2            & 3.97 & 4.40 & 3.36 & 3.58 & 4.52 \\
v3            & 3.98 & 4.45 & 3.29 & 3.64 & 4.53 \\
v4            & 3.89 & 4.49 & 3.18 & 3.46 & 4.43 \\
v5            & 3.91 & 4.43 & 3.10 & 3.57 & 4.53 \\
v6            & 4.00 & \textbf{4.56} & 3.28 & 3.53 & \textbf{4.64} \\
v7 (selected) & \textbf{4.08} & 4.51 & 3.50 & \textbf{3.88} & 4.43 \\
v8            & 4.07 & 4.46 & 3.46 & 3.76 & 4.58 \\
v9            & 3.95 & 4.50 & 3.26 & 3.53 & 4.51 \\
v10           & 3.92 & 4.21 & \textbf{3.56} & 3.70 & 4.24 \\
\midrule
$\Delta$ (v7 -- v0) & +0.27 & +0.15 & +0.53 & +0.56 & $-$0.15 \\
\bottomrule
\end{tabular}
\end{table*}
\section{Additional Qualitative Traces}
\label{app:qualitative-traces}

We report three additional rollouts in which the inner-loop components intercept a baseline failure on the same patient. The first two illustrate the treatment-completion subagent; the third illustrates
the diagnosis-formation subagent, whose broader differential prevents an anchoring
error from cascading into a missed treatment. The trauma case is drawn from a
held-out batch, giving a concrete instance of the same interception on a patient
the proposer never observed during harness search.

\paragraph{Trauma resuscitation completion.}
Patient B10-P07 arrives with an open leg fracture that has also damaged a major
artery, and is in shock from blood loss. Both systems correctly identify the
fracture and the arterial injury and route the patient to the operating room, so
the two traces differ only in how completely the resuscitation is carried out. On the baseline trace, the Main Agent gives antibiotics, pain control, and splinting but omits tetanus prophylaxis, early TXA, and hemostatic resuscitation despite hemorrhagic shock. As a result, the arterial-injury blood loss remains undertreated and the patient's blood pressure continues to fall before operative transfer. On the \OURAGENT{} trace, the
treatment-completion subagent cross-references the open-fracture trauma regimen in
the skills library and flags the three gaps, which the Main Agent then closes,
completing five of the six required interventions before transfer to the
operating room.

\paragraph{Sepsis antibiotic completion.}
Patient B6-P04 presents with abdominal pain, fever, tachycardia, marked leukocytosis, pyuria, and acute kidney injury on a background of advanced bladder cancer, consistent with probable complicated urinary-source sepsis. On the baseline trace, the Main Agent treats the malignancy and metabolic abnormalities but admits the patient without antimicrobial therapy. On the \OURAGENT{} trace, the treatment-completion subagent flags the absent empiric antibiotic, and the Main Agent initiates broad-spectrum coverage before admission.

\paragraph{Atypical NSTEMI recognition.}
Patient B4-P03 presents with altered mental status, abnormal kidney labs, and a history of cardiovascular disease. On the baseline trace, the Main Agent anchored on the abnormal kidney values, attributed the altered mental status to renal/metabolic encephalopathy, and admitted the patient to a general medical ward without ordering an ECG or troponin. Because a cardiac cause was never considered, the NSTEMI remained undetected. On the \OURAGENT{} trace, the diagnosis-formation subagent kept ACS in the differential despite the atypical presentation. The Main Agent ordered an ECG and serial troponins, which showed ischemic ECG changes and elevated/dynamic troponin, supporting NSTEMI. The patient then received cardiology-directed ACS treatment and was admitted to a monitored cardiology unit.
\end{document}